\documentclass[10pt,twocolumn,letterpaper]{article}

\usepackage{iccv}
\usepackage{times}
\usepackage{epsfig}
\usepackage{graphicx}
\usepackage{amsmath}
\usepackage{amssymb}
\usepackage{cite}

\usepackage[breaklinks=true,bookmarks=false]{hyperref}

\iccvfinalcopy 

\def\iccvPaperID{915} 

\begin{document}

\title{An Adaptive Descriptor Design for Object Recognition in the Wild}

\author{Zhenyu Guo, Z. Jane Wang\\
Dept. of ECE, University of British Columbia\\
2332 Main Mall\\
Vancouver, BC Canada V6T 1Z4\\
{\tt\small \{zhenyug, zjanew\}@ece.ubc.ca}
}

\maketitle

\begin{abstract}
   Digital images nowadays have various styles of appearance, in the aspects of color tones, contrast, vignetting, and etc. These `picture styles' are directly related to the scene radiance, image pipeline of the camera, and post processing functions. Due to the complexity and nonlinearity of these causes, popular gradient-based image descriptors won't be invariant to different picture styles, which will decline the performance of object recognition. Given that images shared online or created by individual users are taken with a wide range of devices and may be processed by various post processing functions, to find a robust object recognition system is useful and challenging. In this paper, we present the first study on the influence of picture styles for object recognition, and propose an adaptive approach based on the kernel view of gradient descriptors and multiple kernel learning, without estimating or specifying the styles of images used in training and testing. We conduct experiments on Domain Adaptation data set and Oxford Flower data set. The experiments also include several variants of the flower data set by processing the images with popular photo effects. The results demonstrate that our proposed method improve from standard descriptors in all cases.
   
\end{abstract}

\section{Introduction\label{sec: intro}}
\begin{figure}[t]

    \begin{minipage}[r]{0.45\linewidth}
        \centering
        \includegraphics[width = 1\linewidth] {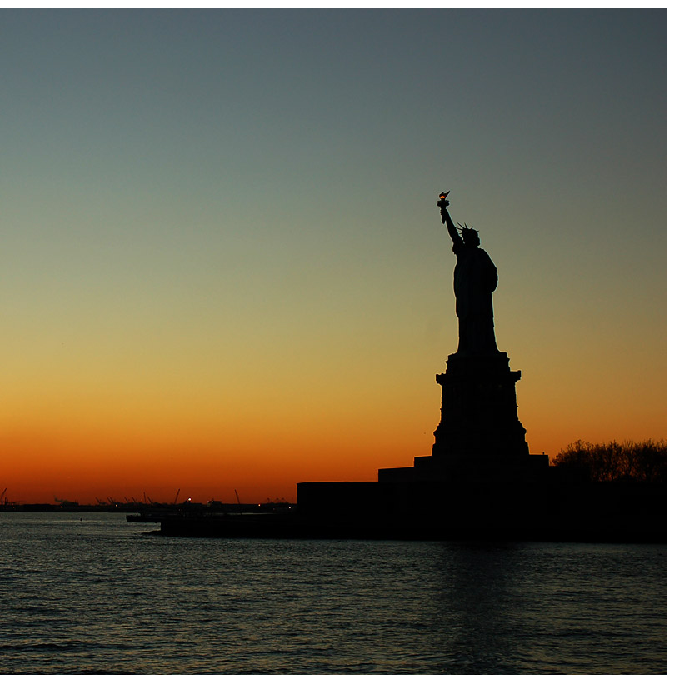}
        (a)
    \end{minipage}
    \hfill
    \begin{minipage}[/]{0.45\linewidth}
        \centering
        \includegraphics[width= 1\linewidth] {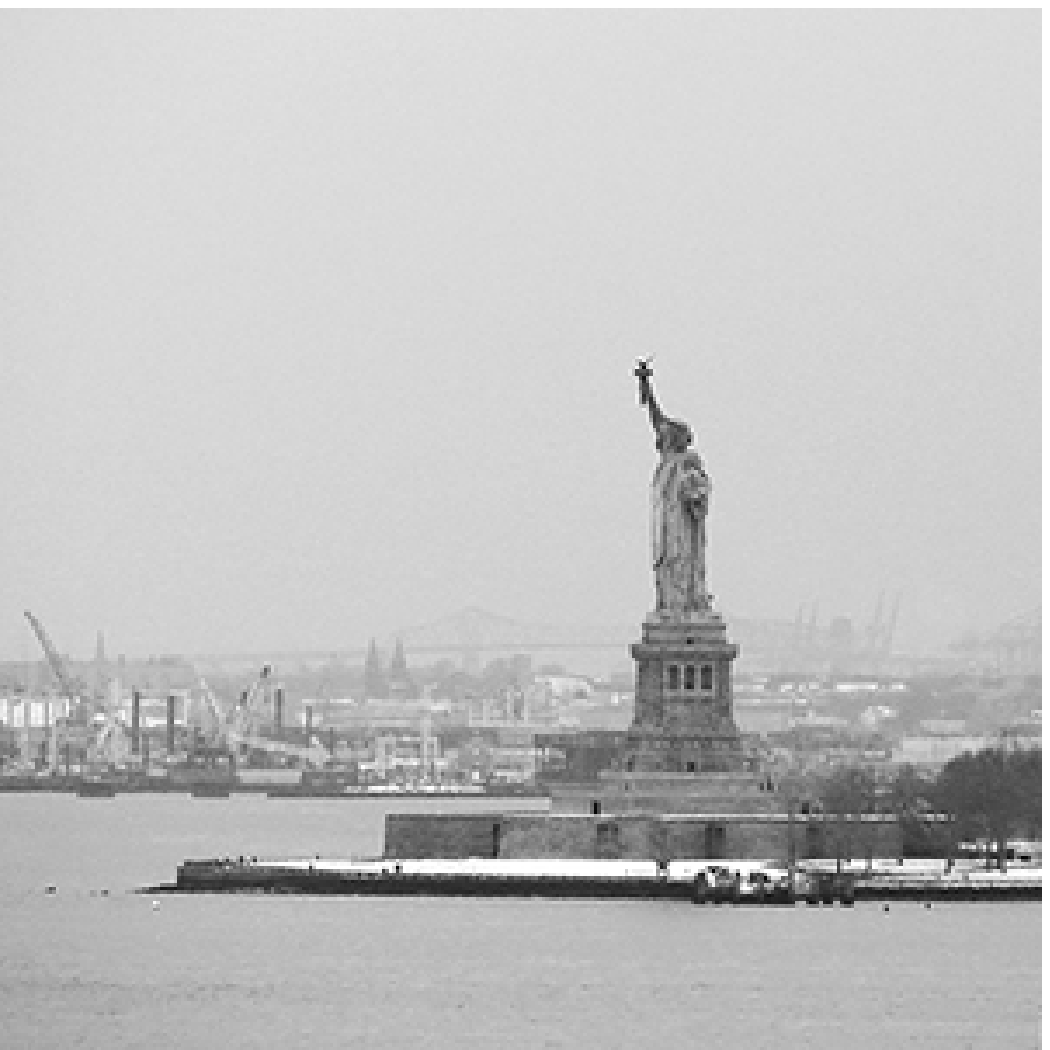}
       (b)
    \end{minipage}

       \begin{minipage}[r]{0.45\linewidth}
        \centering
        \includegraphics[width = 1\linewidth] {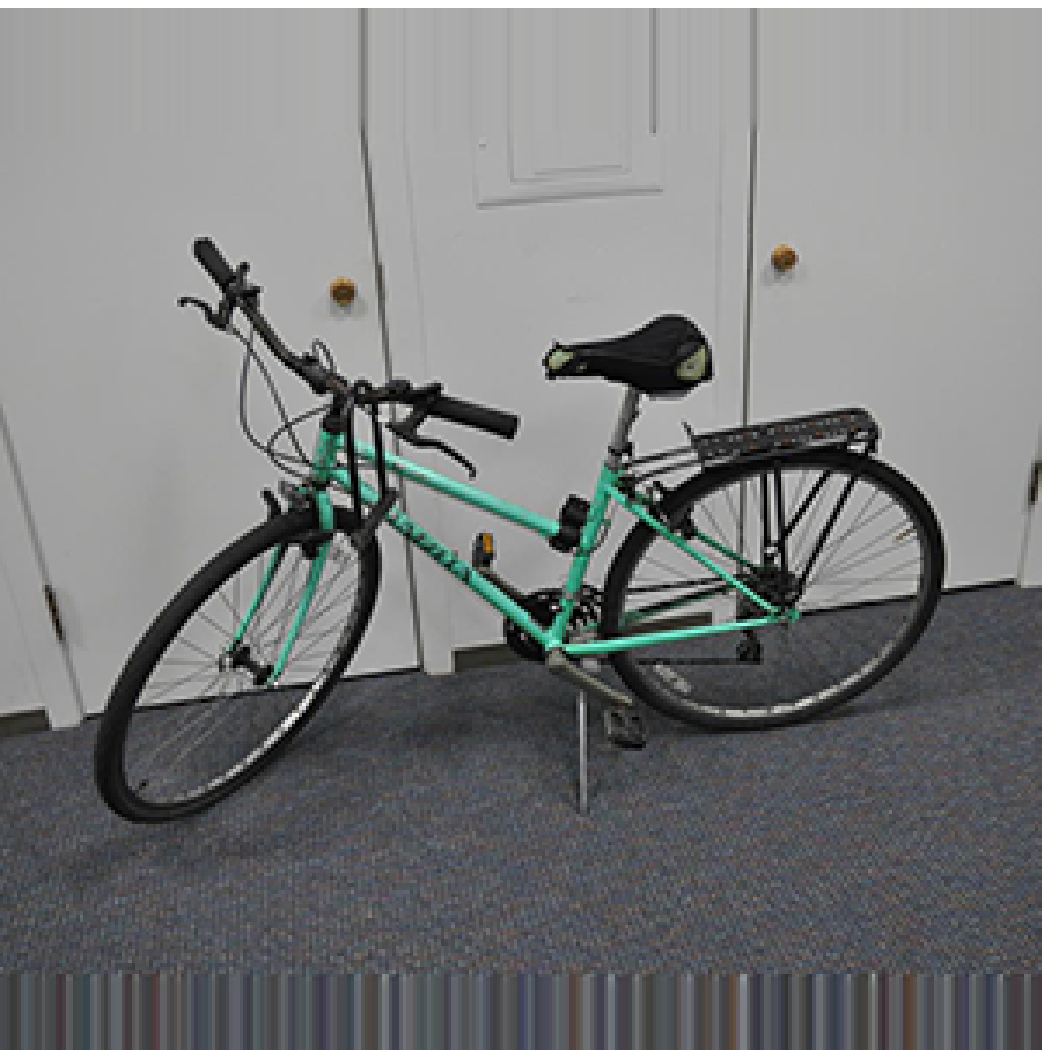}
      (c)
    \end{minipage}
    \hfill
    \begin{minipage}[/]{0.45\linewidth}
        \centering
        \includegraphics[width= 1\linewidth] {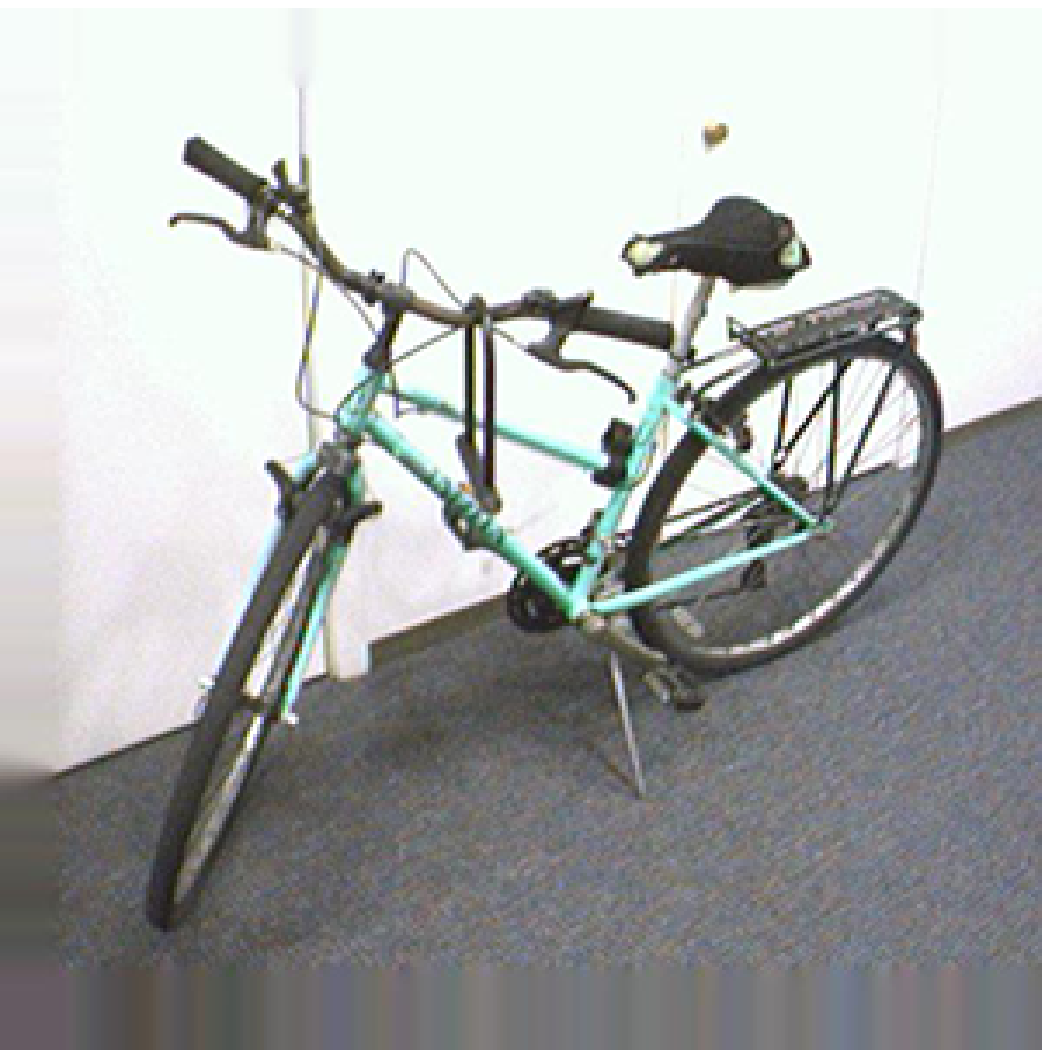}
       (d)
    \end{minipage}

    \begin{minipage}[r]{0.45\linewidth}
        \centering
        \includegraphics[width = 1\linewidth] {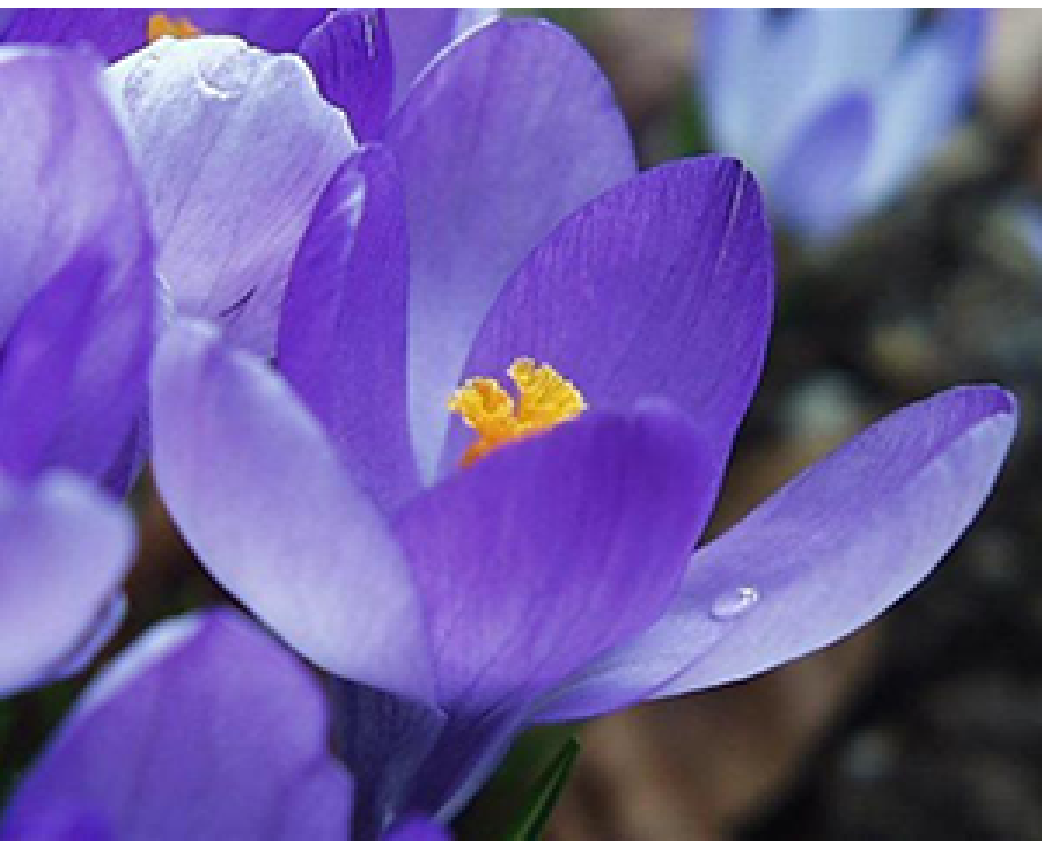}
      (e)
    \end{minipage}
    \hfill
    \begin{minipage}[/]{0.45\linewidth}
        \centering
        \includegraphics[width= 1\linewidth] {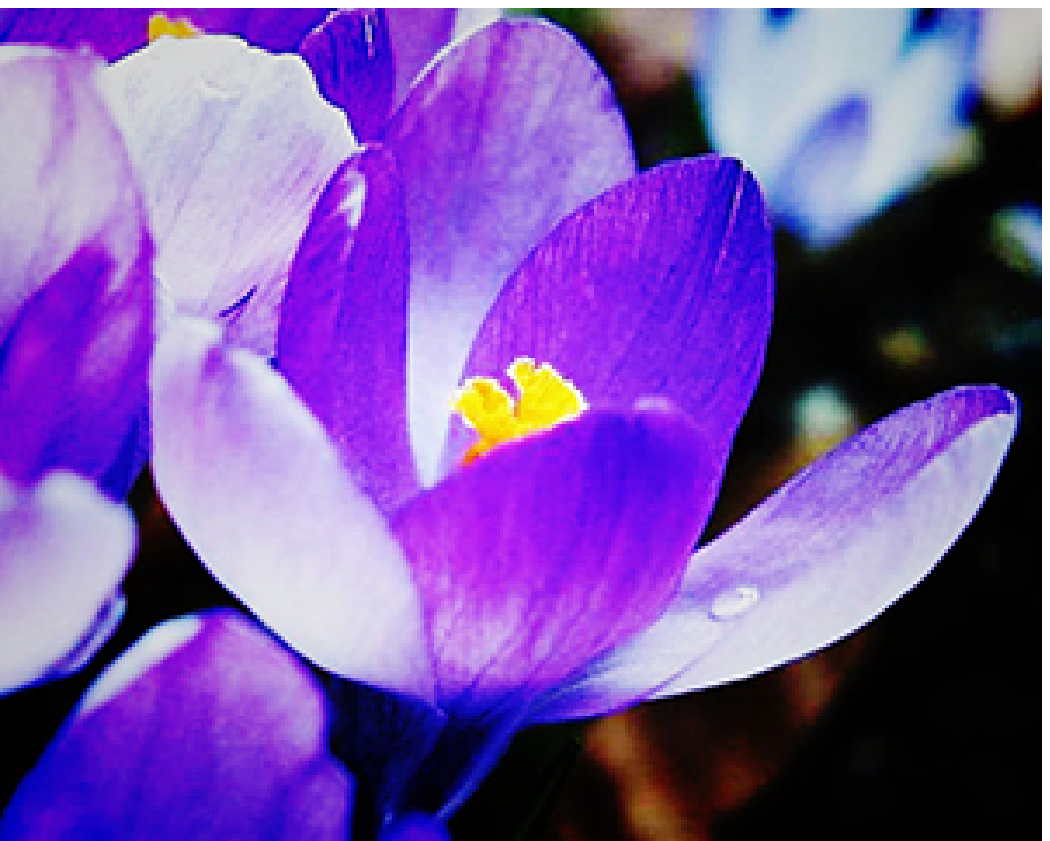}
       (f)
    \end{minipage}

\caption{We show 3 pairs of images with different picture styles about the same objects. The difference between (a) and (b) are mainly caused by different scene radiances(illumination condition). (c) and (d) are of the same object and taken under the same condition by a digital SLR and a webcam respectively, which represent two different image pipelines. (f) is an image obtained by applying Instagram\texttrademark {\it lomo-fi} effect filter to image (e), which is one kind of post processing.}
\label{fig: pic_style}
\end{figure}

\begin{figure*}
\begin{center}
\includegraphics[width = 0.9\linewidth] {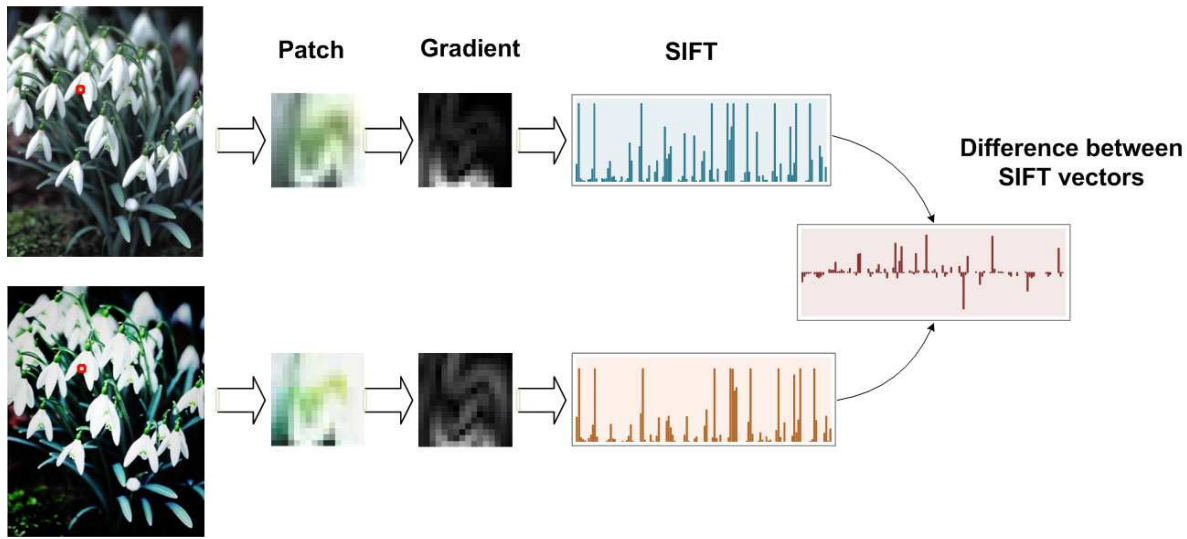}

\end{center}
   \caption{In the upper left is an original image from Oxford Flower data set. In the lower left is the {\it lomo-fi} version of the image. We select two frames of the two images at the same location (red boxes), and show the pixel patch, gradient, and SIFT descriptor for each of them. We plot the difference between two descriptors in the right.}
\label{fig: pic_des}
\end{figure*}
Digital images may vary in the aspects of color tones, contrast, clarity, vignetting, and etc.. We refer such characteristics of digital images as {\bf picture styles} in general. With the popularity of photo editing and sharing services such as Instagram, Facebook and Flickr that are available on mobile devices, most of the digital images generated by users nowadays are captured with a wide range of devices (e.g., smart phones and digital slrs) and processed by different pixel editing functions (e.g., ``lomo-fi'' and ``lord-kelvin'' available in Instagram) to get distinct picture styles with strong personal artistic expressions. Recall that the goal of object recognition research is to recognize natural scenes\cite{lazebnik06spm}, daily objects\cite{feifei06one}, or fine-grained species\cite{nilsback06flower, wah11bird} based on digital images, it is natural to extend the scope of object recognition from standard laboratory images to photos in the wild for daily use. Although there are a large number of picture styles out there, the causes can be separated into 3 categories: (1) scene radiance, (2) image pipeline, and (3) post processing. In Fig. \ref{fig: pic_style} we show several pairs of images about the same objects, with different picture styles.

To show the connection between image descriptors with the picture styles, we take an image from the Oxford Flower data set and process it with a popular {\it Instagram} effect filter: {\it lomo-fi}. We select two patches at the same locations for these two images respectively, and compute the gradients and SIFT descriptors based on the patches, which are shown in Fig. \ref{fig: pic_des}. Although these two image patches are almost same except the color tones, we found that the resulting SIFT descriptors differ with each other in 33\%, which probably will make them be quantized into two dictionary words in the bag-of-word model. The difference is already this much for two images those are almost identical in content, it is reasonable to infer that the difference could be larger for two images with different picture styles within one object class. Therefore, when images used during training and testing don't have similar picture styles, the accuracy of the object recognition will drop. Among the previous work, only {\it Domain Adaptation} (DA) considered one situation\cite{saenko10da} where part of the images are taken by a Digital SLR and the rest are taken by a webcam under similar conditions (e.g. (c) and (d) in Fig. \ref{fig: pic_style}). And they assume images used in training are taken by different device from the images used in testing.

Although the DA gave some solutions to this special case by considering the two sets of images as two domains, in their algorithms the domain-ship of an image has to be specified. However, in the real world applications, images collected from Internet have no ``domain labels'', and the training /testing sets are always mixtures of all kinds of images with various picture styles. Furthermore, much more picture styles are created by users (e.g. Instagram users or iphone camera app users) besides the ones caused by different camera responses. Therefore, in a more general assumption than DA, to find a robust object recognition algorithm becomes useful and challenging, which should overcome the difficulties introduced by different picture styles without knowing the style information.

In this paper, we study this general problem with a focus on descriptor design. Existing approaches usually ignore the changes of picture styles when computing the standard descriptors, then they try to reduce the influences of style changes through operations in the corresponding feature space. Such indirect methods are limited by the feature space and always require the style information for each of the images (e.g. the domain-ship in DA research). In this paper, we take the direct way. Suppose picture styles of all the images in a data set form a point in a certain space, then the original data set is corresponding to a point $A$ in this space. We define a function $g: [0, 255] \rightarrow [0, 255]$ which can be applied to all the images in the original data set and project point $A$ to $B$, where $B$ corresponds to the new data set of all the processed images and can be denoted as $B=g(A)$. Basically $g(\cdot)$ is a pixel-level editing function which works indifferently for all images. Since the picture styles of an image affect the object recognition performance through the descriptors, we assume there is an optimal $g^*$ that maps $A$ to $B^*$ where the set of training and testing images gives best recognition accuracy. The searching for $g^*$ could be difficult since there is no clear connections between a general function $g$ with the empirical risk of the classifier used in object recognition. However, by defining $g$ based on a convex combination of several base functions, in this paper, we link the pixel editing function with image descriptors and kernels. And we also propose an adaptive descriptor design based on kernel learning to achieve the equivalent objective. We derive the method based on kernel descriptors\cite{bo10kdes}, but the ultimate algorithm can be extended to existing standard descriptors as a framework in general. In the following, we discuss some related works in Section \ref{sec: related}. Then we revisit the kernel descriptors in Section \ref{sec: kdes}. We present the proposed method in Section \ref{sec: adapt_des} and Experiments in Section\ref{sec: exp}.


\subsection{Related Works\label{sec: related}}
Domain Adaptation is probably the most related area to our problem. In the data set introduced in\cite{saenko10da}, images from {\it dslr} and {\it webcam} mainly differ in picture styles, which is similar to the focus of this paper. Metric learning based methods \cite{saenko10da, kulis11art} and Grassmann manifold based methods \cite{gong12geodesic, gopalan11domain} were proposed. As we stated, these DA methods cannot solve our proposed problem in general situations, since the domain-ship is unknown and hard to specify for image in the wild.
Works like \cite{grossberg04modeling, xiong12pixels, kim12new} estimate the model of image pipelines, but such estimations are difficult to get and has no clear relationship with the descriptors and recognition accuracy. 
In the area of key point matching, several robust descriptors are proposed, such as DAISY\cite{tola10daisy}, GIH\cite{ling05deformation} and DaLI\cite{moreno11deformation}. Descriptor learning methods\cite{winder07learning, winder09picking, simonyan12descriptor} are also invented to determine the parameters in descriptor computation by optimization. All of these methods are designed for key point matching between image pairs. The different goal leads to descriptors that are not suitable for object recognition, since they are too discriminative to tolerant the with-in class variance for object categories.

\section{Kernel Descriptor Revisit\label{sec: kdes}}
The kernel descriptor (KDES) is proposed by Bo et. al. in \cite{bo10kdes}, which gives a unified framework and parametric form for local image descriptors. Let $z$ denote a pixel at coordinate $z$, $m(z)$ denote the magnitude of image gradient at pixel $z$, and $\theta(z)$ denote the orientation of image gradient. And $m(z)$ and $\theta(z)$ are the normalized by the average values of one patch contains $z$ into $\tilde m(z)$ and $\tilde \theta(z)$.
According to this kernel view, the gradient matching kernel between two image patches $P$ and $Q$ can be described as
\begin{equation}\label{eq: kdes}
k_{grad}(P,Q) = \sum_{z\in P}\sum_{z'\in Q}\tilde m(z)\tilde m(z')k_o(\tilde \theta(z), \tilde \theta(z'))k_p(z,z'),
\end{equation}
where $k_p(z,z')= exp(-\gamma_p ||z-z'||^2)$ is a Gaussian position kernel and $k_o(\tilde \theta(z), \tilde \theta(z')) = exp (-\gamma_o||\tilde \theta(z)-\tilde \theta(z')||^2$ is a Gaussian kernel over gradient orientations. And $\tilde m(z) = m(z) / \sqrt{\sum_{z\in P} m(z)^2 + \epsilon_g}$, where $\epsilon_g$ is a small number. Orientation is normalized as $\tilde \theta(z) = [sin(\theta(z))cos(\theta(z))]$. To build compact feature vectors from these kernels for efficient computation, \cite{bo10kdes} presented a sufficient finite-dimensional approximation to obtain finite-dimensioned feature vectors and reduce the dimension by kernel principal component analysis, which provides a close form for the descriptor vector $F_{grad}(P)$ of patch $P$ such that $k_{grad}(P,Q) = F_{grad}(P)^TF_{grad}(Q)$. And Bo et. al. \cite{bo10kdes} also showed that gradient based descriptor like SIFT\cite{lowe04sift}, SURF\cite{bay08surf}, and HoG\cite{dalal05hog} are special cases under this kernel view framework.

For the image-level descriptors, Bo and Sminchisescu\cite{bo09efficient} presented Efficient Match Kernels (EMK) which provided a general kernel view of matching between two images as two sets of local descriptors. And they demonstrated that Bag-of-Word (BoW) model and Spatial Pyramid Matching are two special cases under this framework. Let $X$ and $Y$ denote two sets of local descriptors for image $I_x$ and $I_y$ respectively. $x\in X$ is a descriptor vector computed from patch $P_x$ in image $I_x$. When applying EMK on top of gradient KDES, we get the image level kernel as 
\begin{equation}\label{eq: emk}
\begin{array}{lcl}
K_{emk}(I_x, I_y) &=& \frac{1}{|X||Y|}\sum_{x\in X}\sum_{y\in Y}k(x,y)\\
&=& \frac{1}{|X||Y|}\sum_{x\in X}\sum_{y\in Y}k_{grad}(P_x,P_y),
\end{array}
\end{equation}
where $|\cdot|$ is the cardinality of a set. \cite{bo09efficient} gave a close form compact approximation of the feature vector such that $K_{emk}(I_x, I_y)=\Phi(I_x)^T\Phi(I_y)$, which makes the matching kernel can be used in real applications with efficient computation and storage.

\section{Proposed Method\label{sec: adapt_des}}

\subsection{Kernel Descriptor with Editing Functions\label{sec: kdes_edit}}
As stated in Introduction, we want to apply a pixel editing function $g$ to images used for object recognition. In this section, we will give the relationship between pixels and descriptors under this function $g$. Take $g(u(z)) = a_1u(z) +a_2u(z)^2$ as an example, where $a_1, a_2$ are non-negative, $z$ is a pixel from image patch $P$, and $u(z)$ is the pixel value at position $z$. Let $g(P)$ denote the new patch after applying $g$ on the pixels of $P$. Then the image gradient at $z$ now becomes
\begin{equation}
\begin{array}{lcl}
\triangledown g(u(z))& = &g'|_{u(z)}\triangledown u(z)\\
&=& (a_1+2a_2u(z))\triangledown u(z),
\end{array}
\end{equation}
where $a_1+2a_2u(z)$ is a scalar and $\triangledown u(z)$ is a vector. Let $m_g(z)$ and $\theta_g(z)$ be the magnitude and orientation of gradient at $z$ of patch $g(P)$, and $m(z)$ and $\theta(z)$ be corresponding values to patch $P$. Under the assumption that $a_1, a_2 \geq 0$, we have 
\begin{equation}\label{eq: gkdes}
\begin{array}{lcl}
m_g(z)& = &||\triangledown g(u(z))|| \\
& = & ||(a_1+2a_2u(z))\triangledown u(z)||\\
& = & a_1||\triangledown u(z)|| + a_2||\triangledown u(z)^2||\\
\end{array}
\end{equation}
It is clear that $\theta_g(z) = \theta(z)$, therefore $\tilde \theta(z) = \tilde \theta_g(z)$, which means the orientation is invariant to the pixel editing functions applied to the image patch. Notice that the magnitude used in gradient match kernel are normalized base on local patches, which is very important to make the contextual information comparable for different patches. Let $m_2(z)$ denote $||\triangledown u(z)^2||$ for convenience. To retain the simple convex combination form of $m_g(z)$, we propose a new locally normalization
\begin{equation}\label{eq: newnorm}
\hat m_g(z) = a_1 \tilde m(z) + a_2 \tilde m_2(z),
\end{equation}
where $\hat m_g(z)$ denotes the new normalized magnitude of $m_g(z)$, and $\tilde m(z)$ and $\tilde m_2(z)$ are normalized by $l_2$ norm as mentioned in Section \ref{sec: kdes}. It is clear that the $\hat m_g(z)$ is also locally normalized and still comparable among different patches. Since the goal of our method is object recognition, any proper locally normalization method is acceptable.

Now given two image patches $g(P)$ and $g(Q)$, which are obtained by applying editing function $g$ to patches $P$ and $Q$, we derive the gradient match kernel between them as following
\begin{equation}\label{eq: sumkernel}
\begin{array}{l}
\hat k_{grad}(g(P), g(Q)) \\
=\sum_{z\in g(P)}\sum_{z'\in g(Q)} \hat m_g(z) \hat m_g(z')k_o(\tilde \theta_g(z), \tilde \theta_g(z'))k_p(z,z')\\
 =  \sum_{z\in P}\sum_{z'\in Q}(a_1 \tilde m(z)+a_2 \tilde m_2(z))(a_1 \tilde m(z') + a_2 \tilde m_2(z'))\\
 k_o(\tilde \theta(z), \tilde \theta(z'))k_p(z,z')\\
 =  a_1 a_1\sum_{z\in P}\sum_{z'\in Q}\tilde m(z)\tilde m(z') k_o(\tilde \theta(z), \tilde \theta(z'))k_p(z,z')\\
 +a_1a_2 \sum_{z\in P}\sum_{z'\in Q}\tilde m(z)\tilde m_2(z')k_o(\tilde \theta(z), \tilde \theta(z'))k_p(z,z')\\
 + a_2a_1  \sum_{z\in P}\sum_{z'\in Q}\tilde m_2(z) \tilde m(z)k_o(\tilde \theta(z), \tilde \theta(z'))k_p(z,z')\\
 +a_2a_2  \sum_{z\in P}\sum_{z'\in Q}\tilde m_2(z)\tilde m(z')k_o(\tilde \theta(z), \tilde \theta(z'))k_p(z,z')\\
 = a_1a_1 k_{grad}(P,Q) +a_1a_2 k_{grad}(P,Q^2) \\
 + a_2a_1 k_{grad}(P^2,Q) + a_2a_2 k_{grad}(P^2,Q^2),
 \end{array}
\end{equation}
where $P^2$ and $Q^2$ denote the patches contain squared pixel values from $P$ and $Q$. And it is worth noting that $\hat k_{grad}$ above is different from the standard $k_{grad}$ in Eq. (\ref{eq: kdes}), since we define a different normalization approach in Eq. (\ref{eq: newnorm}). Let's rewrite $g$ as $g(u(z)) = a_1 g_1(u(z)) + a_2 g_2(u(z))$, where $g_1 (u(z)) = u(z)$ and $g_2(u(z)) = u(z)^2$. Therefore Eq. (\ref{eq: sumkernel}) indicates 
\begin{equation}\label{eq: kernel_algo}
\hat k_{grad}(g(P),g(Q)) = \sum_{i=1}^2\sum_{j=1}^2{a_ia_jk_{grad}(g_i(P),g_j(Q))}.
\end{equation}
As stated in Introduction, we successfully link the pixel editing functions with image descriptors. To see this connection in image-level, we plug Eq. (\ref{eq: kernel_algo}) into Eq. (\ref{eq: emk}) and get the image-level kernel
\begin{equation}\label{eq: im_mkl}
\begin{array}{l}
K_{emk}(g(I_x), g(I_y) = \frac{1}{|X||Y|}\sum_{x\in X}\sum_{y\in Y}\hat k_{grad}(g(P_x),g(P_y)\\
= \frac{1}{|X||Y|}\sum_{x\in X}\sum_{y\in Y}\sum_{i=1}^2\sum_{j=1}^2{a_ia_jk_{grad}(g_i(P_x),g_j(P_y))}\\
=\sum_{i=1}^2\sum_{j=1}^2a_ia_j(\frac{1}{|X||Y|}\sum_{x\in X}\sum_{y\in Y}k_{grad}(g_i(P_x),g_j(P_y)))\\
=\sum_{i=1}^2\sum_{j=1}^2a_ia_j K(g_i(I_x),g_j(I_y))\\
=\sum_{m=1}^4 d_m K_m,
\end{array}
\end{equation}
where $d_m$ and $K_m$ have one-to-one correspondence to $a_ia_j$ and $K(g_i(I_x),g_j(I_y))$, and the order does not matter since they are exchangeable in the summation. We limit $a_i$ to be non-negative when we first defined $g$, then $d_m$'s are also non-negative. In addition, $K_m$'s are positive definite(PD) kernels, which makes $K_{emk}$ here a convex combination of PD kernels and can be used in standard multiple kernel learning. Therefore, we successfully transfer the searching of optimal $g^*$ into learning the optimal kernel weights through Eq. (\ref{eq: im_mkl}). In general, for $N$ based editing functions, there will be $N^2$ base kernels in Eq. (\ref{eq: im_mkl}).



\begin{figure}
\begin{center}
\includegraphics[width = 0.9\linewidth] {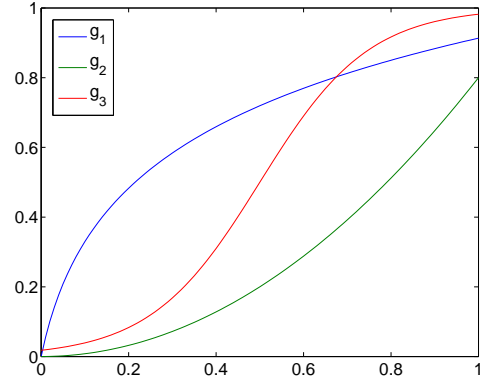}

\end{center}
   \caption{Plots of proposed base functions.}
\label{fig: editfuncs}
\end{figure}

\begin{figure*}
\begin{center}$
\begin{array}{cccc}
\includegraphics[width=0.2 \linewidth]{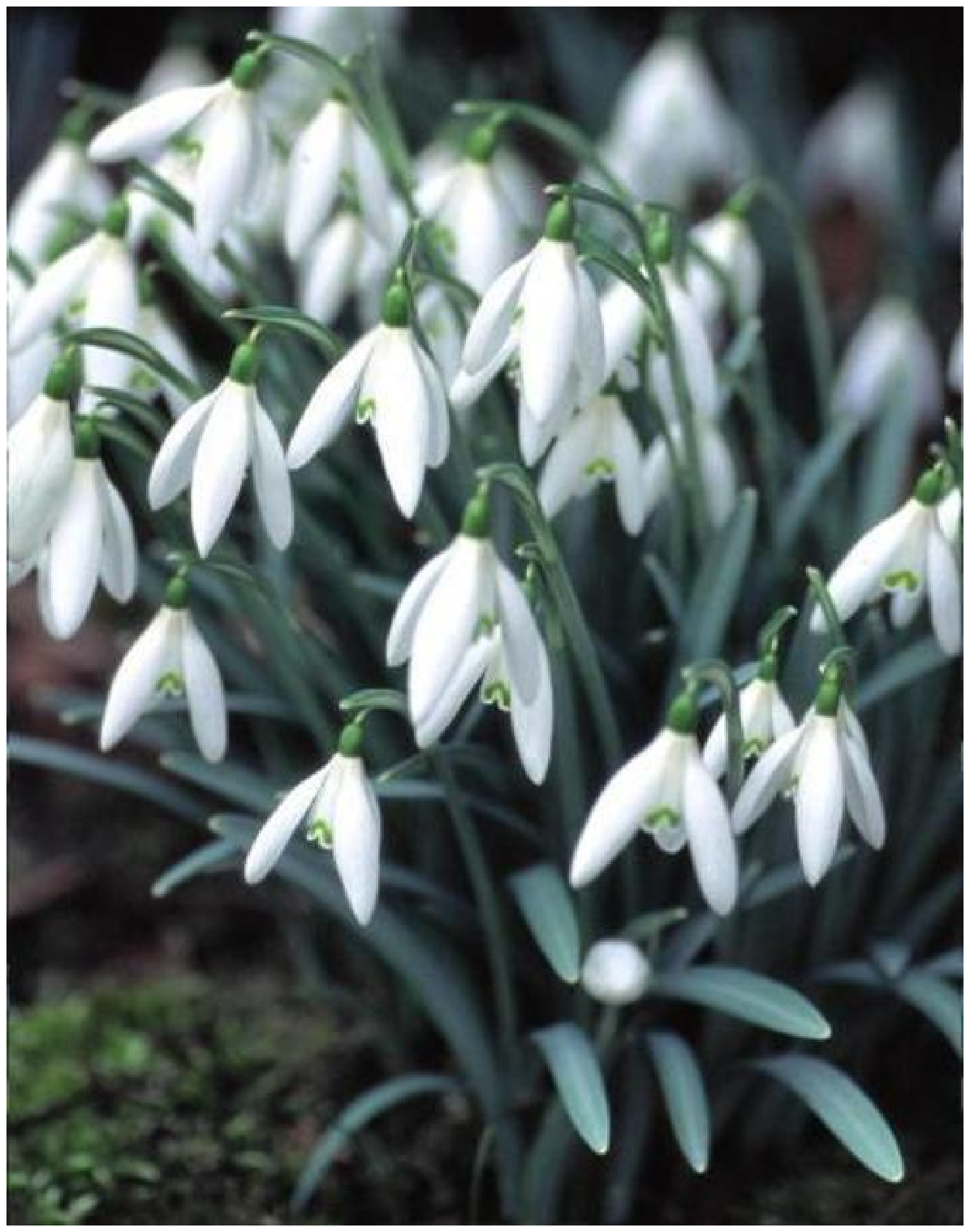} &
\includegraphics[width=0.2\linewidth]{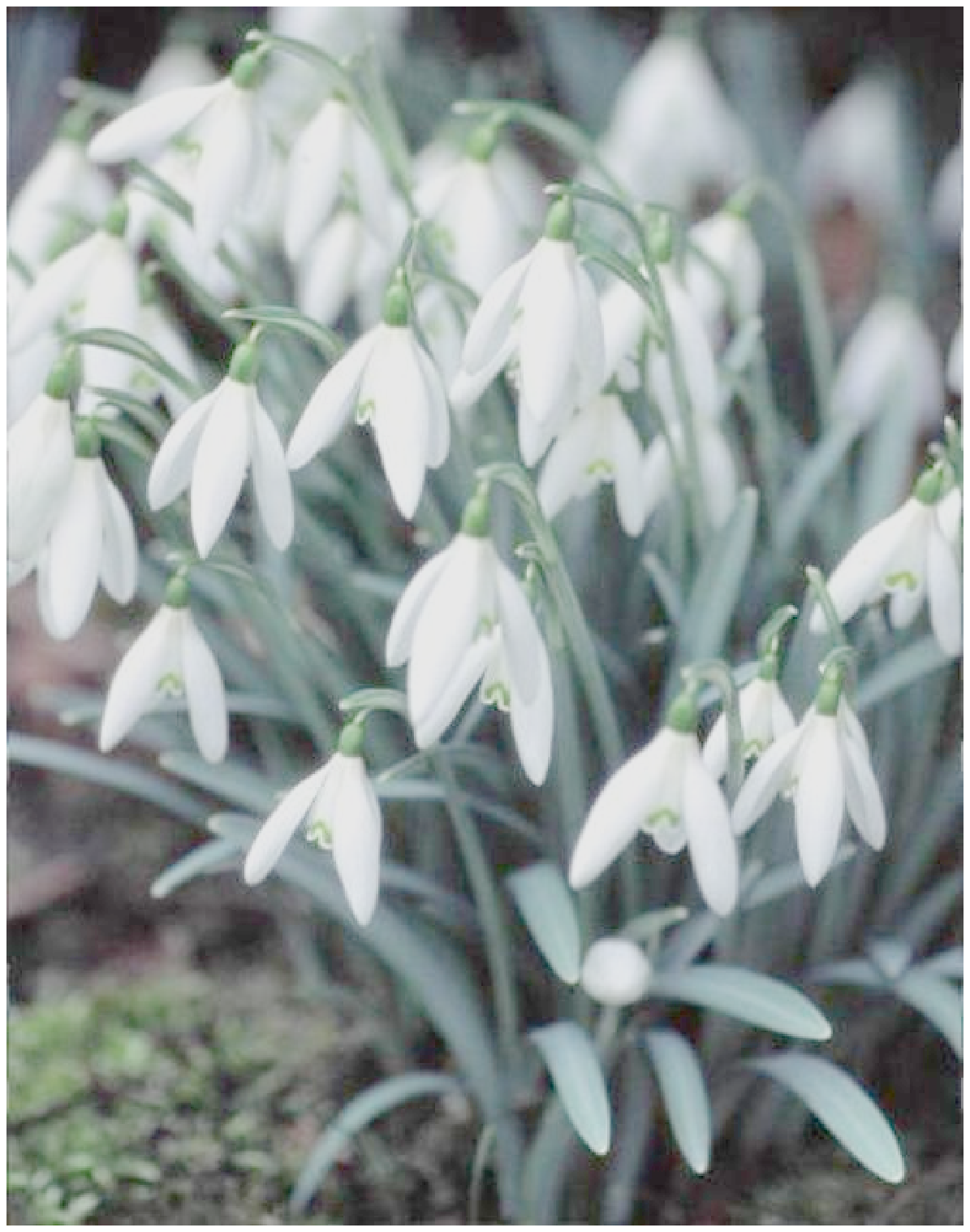} &\includegraphics[width=0.2\linewidth]{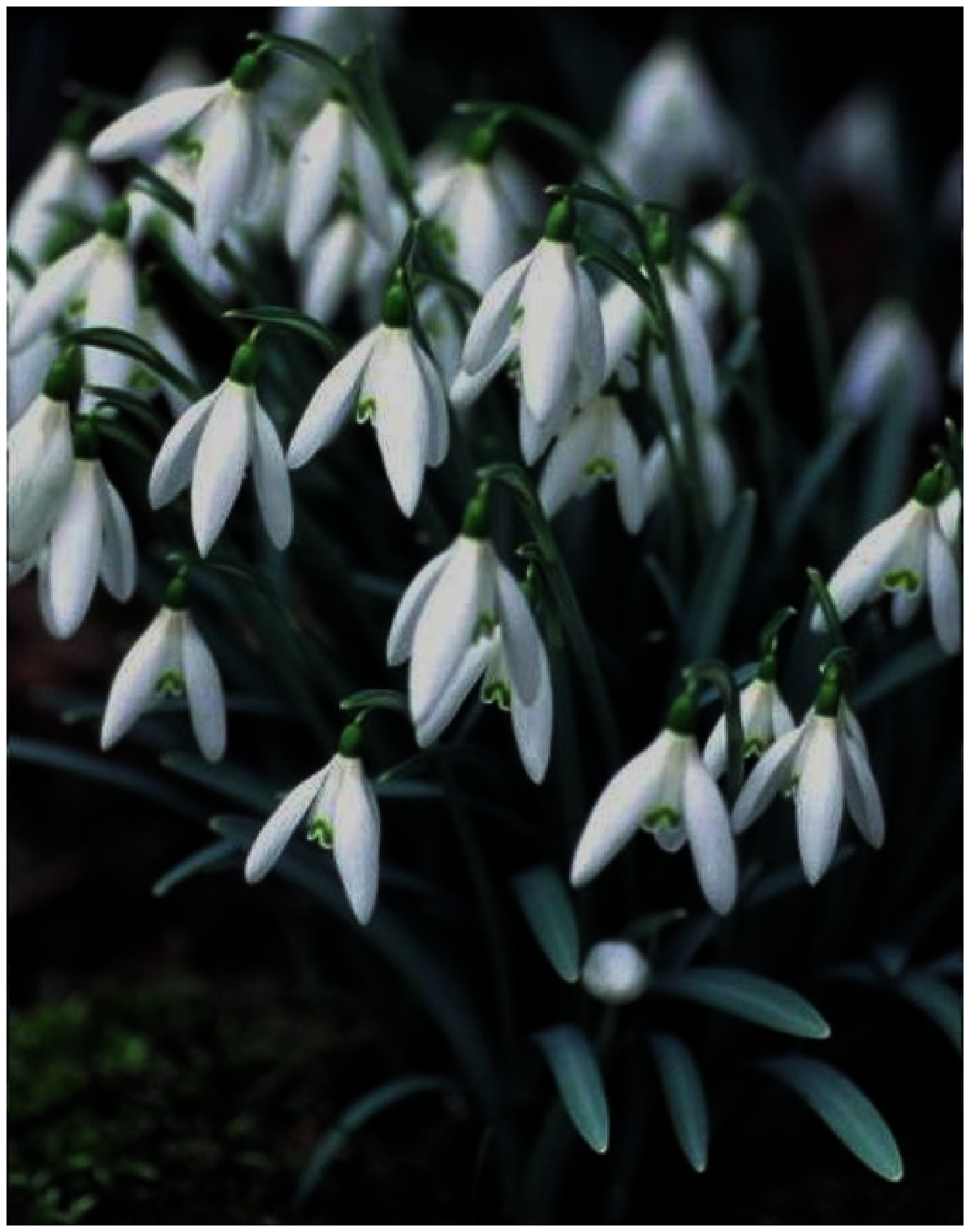} &\includegraphics[width=0.2\linewidth]{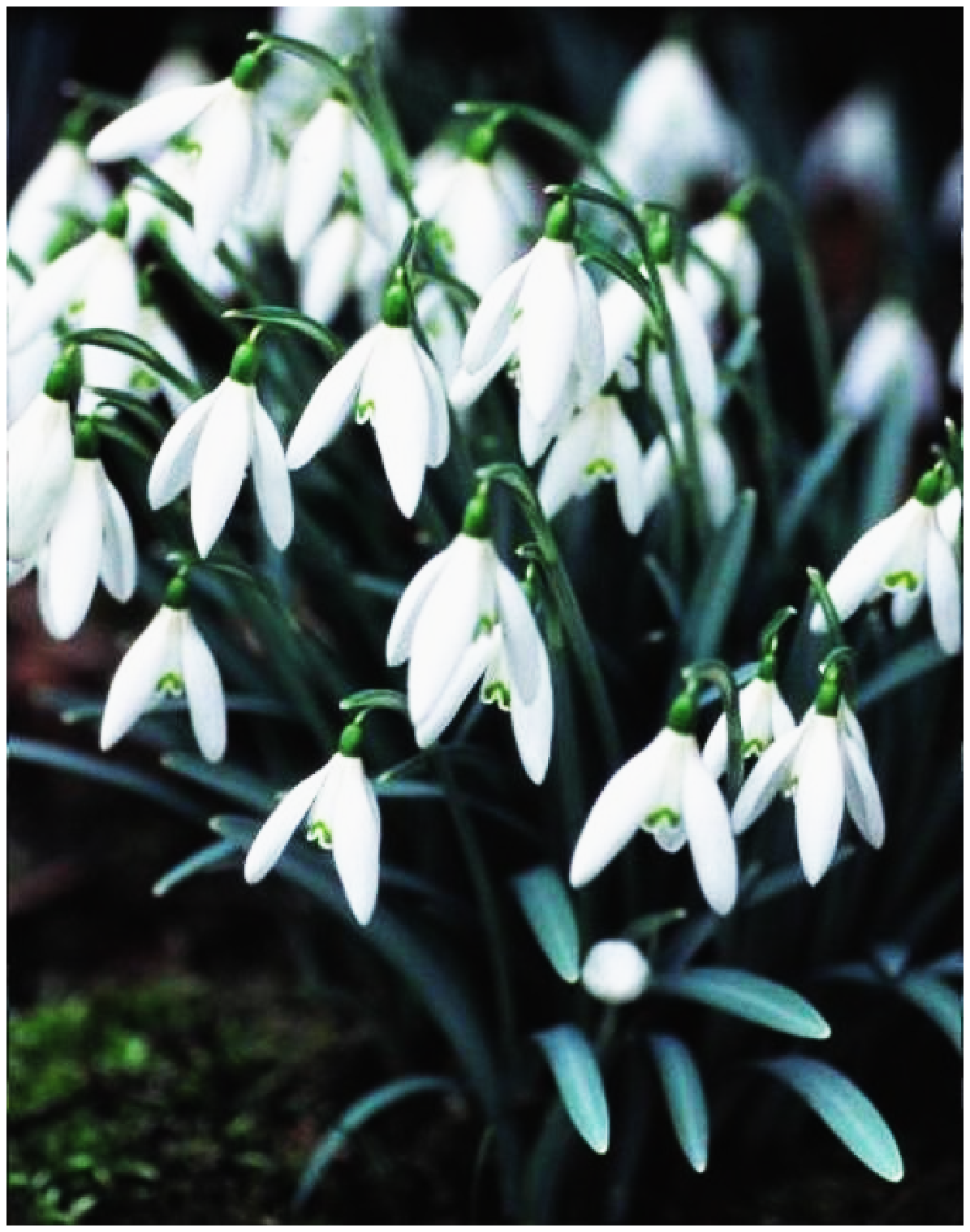}
\end{array}$
\end{center}
\caption{From left to right: original image $I$, $g_1(I)$, $g_2(I)$, and $g_3(I)$.}
\label{fig: effectim}
\end{figure*}
\subsection{Base Editing Functions}
From Section \ref{sec: kdes_edit}, we know that the selection of base functions is as important as learning the parameters. By exploring the editing functions used for photography, we found Gamma correction and the ``S'' curve are two major categories of photo effects. Gamma correction can brighten($\gamma < 1$) or darken ($\gamma >1$) the images, and the ``S'' curve can increase the contrast. Although these popular functions are created for the visually pleasure of photos, we believe that they can also benefit the computation of better descriptors. For example, brightening can bring back more details in the dark part of an image; Darkening can surpass the irrelevant areas of an object image, since most of the images are correctly exposed for the centering object; Higher contrast emphasizes the texture and shapes. Therefore, these three types of functions make good candidates for base functions. However, a power-law function used in Gamma correction contains a free parameter that will be left in the gradient and cannot be combined using simple addition, then a good approximation is desired. And the ``S'' curve doesn't have a standard formulation, we adopt the sigmoid function for our algorithm. The three based functions that we are proposing here are:
\begin {equation}
g_1(x) = 0.3*(\log(2x+0.1)+|\log(0.1)|)
\end{equation}
\begin {equation}
g_2(x) = 0.8x^2
\end{equation}
\begin {equation}
g_3(x) = \frac{1}{1+e^{-8(x-0.5)}},
\end{equation}
where $x$ takes a value from $[0,1]$ as a scaled image pixel. We show the plots of these functions in Fig. \ref{fig: editfuncs}. From the plots we can see that $g_1$ could serves as a brightening function, similar to gamma correction when $\gamma<1$. $g_2$ has a shape like gamma correction when $\gamma=2$, which could be used as darkening. And $g_3$ has a ``S'' shape that increase the contrast by brightening the brights and darkening the darks. The effects of these functions can be seen from Fig.\ref{fig: effectim}.

\subsection{Learning of the Parameters\label{sec: la}}
After defining the base functions, the next question is how to estimate the weighting coefficients for object recognition. According to Eq. (\ref{eq: im_mkl}), the image-level kernel can be decomposed as a convex combination of several base kernels. We adopt General Multiple Kernel Learning (GMKL)\cite{varma09gmkl} and put non-negative constraints on the kernel coefficients. We also notice that some weighting coefficients are same (e.g. $a_1a_2 = a_2a_1$), but the experiments show that the results are similar with or without this constraint on the weights. And the number of kernels in our algorithm is not large, therefore, we use the standard GMKL with $l2$ norm regularization.


\subsection{Adaptive Descriptor Design\label{sec: add}}
In this section, we summarize the proposed adaptive descriptor design in the following
\begin{enumerate}
\item Process image $I$ from the data set with $\{g_i\}_{i=1}^3$.
\item Compute gradient-based descriptors for $I$ and its 3 variants to get 4 descriptors.
\item Build a codebook using Kmeans by sampling from all the training images and all 4 descriptors of each.
\item Quantize each image from training and testing sets into 4 image-level feature vectors based on the descriptors.
\item For two images, compute linear kernels between any two of their 4 image-level features to get 16 base kernels.
\item Train GMKL on 16 base kernels to get optimal kernel weights and classifiers. 
\end{enumerate}

Our proposed method does not require prior knowledge on picture styles of training or testing images, and the Adaptive Descriptor Design (ADD) can work as a general framework with flexibility. In step 1, other proper functions can be used here as base editing functions, besides the ones we used here. According to the analysis by \cite{bo10kdes}, most of the gradient-based descriptors, such as SIFT\cite{lowe04sift}, SURF\cite{bay08surf} and HoG\cite{dalal05hog}, are special cases of the kernel descriptor(KDES), which all can be used in step 2 to compute descriptors from image patches. In addition, the quantization method used in step 4 can be chose from Bag-of-word, Spatial Pyramid Matching and Efficient Match Kernel, since the former two are special cases of EMK. In other words, our proposed algorithm can be used widely to improve the previous methods which are based on gradient descriptors and SVMs.

We also want to point out that the proposed ADD is a single feature method, although a GMKL is used for estimating the coefficients. Essentially, the final kernel obtained through GMKL is equivalent to a single kernel based on standard descriptors of images processed by optimal $g^*$.

\section{Experiments\label{sec: exp}}
In this section, we describe the details of experiments and report the results for the proposed method to compare with standard gradient-based image descriptors. We conduct object recognition on Domain Adaptation data set and Oxford Flower data set. We also process the images from Oxford Flower data set using several popular photo effects in Instagram\texttrademark \footnote{We use Adobe Photoshop\texttrademark action files created by Daniel Box, which can give similar effects as Instagram\texttrademark.}.

\subsection{Domain Adaptation}
Domain Adaptation data set was introduced by \cite{saenko10da}, where images for the same categories of objects are from different sources (called domains): {\it amazon}, {\it dslr} and {\it webcam}. As we stated in Introduction, the two domains {\it dslr} and {\it webcam} only differ in picture styles which are caused by image pipelines. Applying the proposed ADD algorithm, we adopt KDES + EMK and SURF+BoW two sets of features to demonstrate that ADD can work as a framework to improve the performance of gradient-based descriptor in general. We follow the experimental protocol used in \cite{saenko10da, kulis11art} for semi-supervised domain adaptation. It is worth noting that we didn't use any domain-ship information to specify the picture styles of images, our proposed method could figure out an optimal descriptor automatically based on the training set. 


\begin{table}
\begin{center}
\resizebox{1\linewidth}{!} {
\begin{tabular}{|l|l||c|c|c|}
\hline
{\it Source} & {\it Target}  & standard KDES & ADD\_AK &  ADD\_GMKL\\
\hline\hline
$dslr$ & $webcam$  &  49.30 $\pm$ 1.26 &  50.28 $\pm$ 1.02 &{\bf 54.81} $\pm$ 1.07\\
$webcam$ & $dslr$  &  46.67 $\pm$ 0.80 &  48.17 $\pm$ 0.99 &{\bf 50.33} $\pm$ 0.79 \\
$amazon$ & $dslr$  &  47.43 $\pm$ 2.79 &  48.57 $\pm$ 2.51 &{\bf 53.90} $\pm$ 2.67\\
\hline
\end{tabular}
}
\end{center}
\caption{Experiments on DA data set based on KDES. The average accuracy in \% is reported and the corresponding standard deviation is included.}
\label{tab: da_kdes}
\end{table}

\subsubsection{ADD based on KDES and EMK\label{sec: daadd}}
We extract KDES descriptors of all the images in three domains and create a 1,500-word codebook by applying K-means clustering on a subset of all 4 types (original + 3 variants for each images) of descriptors from {\it amazon} domain. And then this codebook is used to quantize 4 types of descriptors of all 3 domains of images using EMK. After obtaining the 16 linear kernels by computing the inner product of every two types of descriptors between two given images, we conduct object recognition experiments using SVMs for: the standard KDES, averaging kernel of this 16 kernels (AK), and GMKL based on 16 kernels. We show the results in Table\ref{tab: da_kdes}, from which we can see that the proposed Adaptive Descriptor Design outperforms the standard KDES in all cases for both averaging kernel and an optimal kernel learned by GMKL. Particularly, the ADD\_GMKL method improved from the standard KDES by all most 6\% in all cases, which is close to the improvements obtained by domain adaptation methods\cite{saenko10da, kulis11art, gong12geodesic, jhuo12robust}, where domain-ship information is used. In addition, we would like to point out that our proposed ADD is actually a {\bf single feature} method since the learned kernel weights can be seen as the combination coefficients for the optimal pixel editing function $g^*$, and the whole process is equivalent to using a single KDES descriptor abstracted from $g^*(I)$ for a given image $I$.

\begin{table}
\begin{center}
\resizebox{1\linewidth}{!} {
\begin{tabular}{|l|l||c|c|c|}
\hline
{\it Source} & {\it Target}  & standard SURF & ADD\_AK & ADD\_GMKL\\
\hline\hline
$dslr$ & $webcam$  &  37.05 $\pm$ 1.72 &  {\bf 41.61} $\pm$ 1.05 &{\bf 42.00} $\pm$ 1.16\\
$webcam$ & $dslr$  &  30.09 $\pm$ 0.81 &  {\bf 36.57} $\pm$ 0.75 &{\bf 36.45} $\pm$ 0.49 \\
$amazon$ & $dslr$  &  34.49 $\pm$ 1.30 &  {\bf 40.62} $\pm$ 1.59 & 36.19 $\pm$ 2.04\\
\hline
\end{tabular}
}
\end{center}
\caption{Experiments on DA data set based on SURF. The average accuracy in \% is reported and the corresponding standard deviation is included.}
\label{tab: da_surf}
\end{table}

\subsubsection{ADD based on SURF and BoW}
To show the generalization ability of ADD, we follow previous methods\cite{saenko10da, kulis11art, gong12geodesic, jhuo12robust} to extract standard SURF descriptors from the original and 3 variants of each image, then a 800-word codebook is created from {\it amazon} domain. All the images in 3 domains are quantized by this codebook using Vector-quantization to get Bag-of-Word features. After obtain 16 linear kernels, we also conduct experiments using standard KDES, averaging kernel, and an optimal kernel learned by GMKL. We report the results in Table \ref{tab: da_surf}. The proposed ADD methods also outperform the standard SURF descriptor in all cases. However, in this experiment, the averaging kernel gives better results than GMKL learned kernel in some of the scenario. We attribute the worse performance of GMKL based ADD to the lack of training, since the SURF descriptors are sparsely extracted from images and only 11 (8 from source domain and 3 from target domain) training images per category are used. But the results for ADD\_AK and ADD\_GMKL are sufficient to show that the proposed Adaptive Descriptor Design can be applied on top of gradient-based descriptors widely, for different tasks.
\subsection{Object Recognition on Oxford Flower with Photo Effects}
\begin{figure*}
\begin{center}$
\begin{array}{cccc}
\includegraphics[width=0.23 \linewidth]{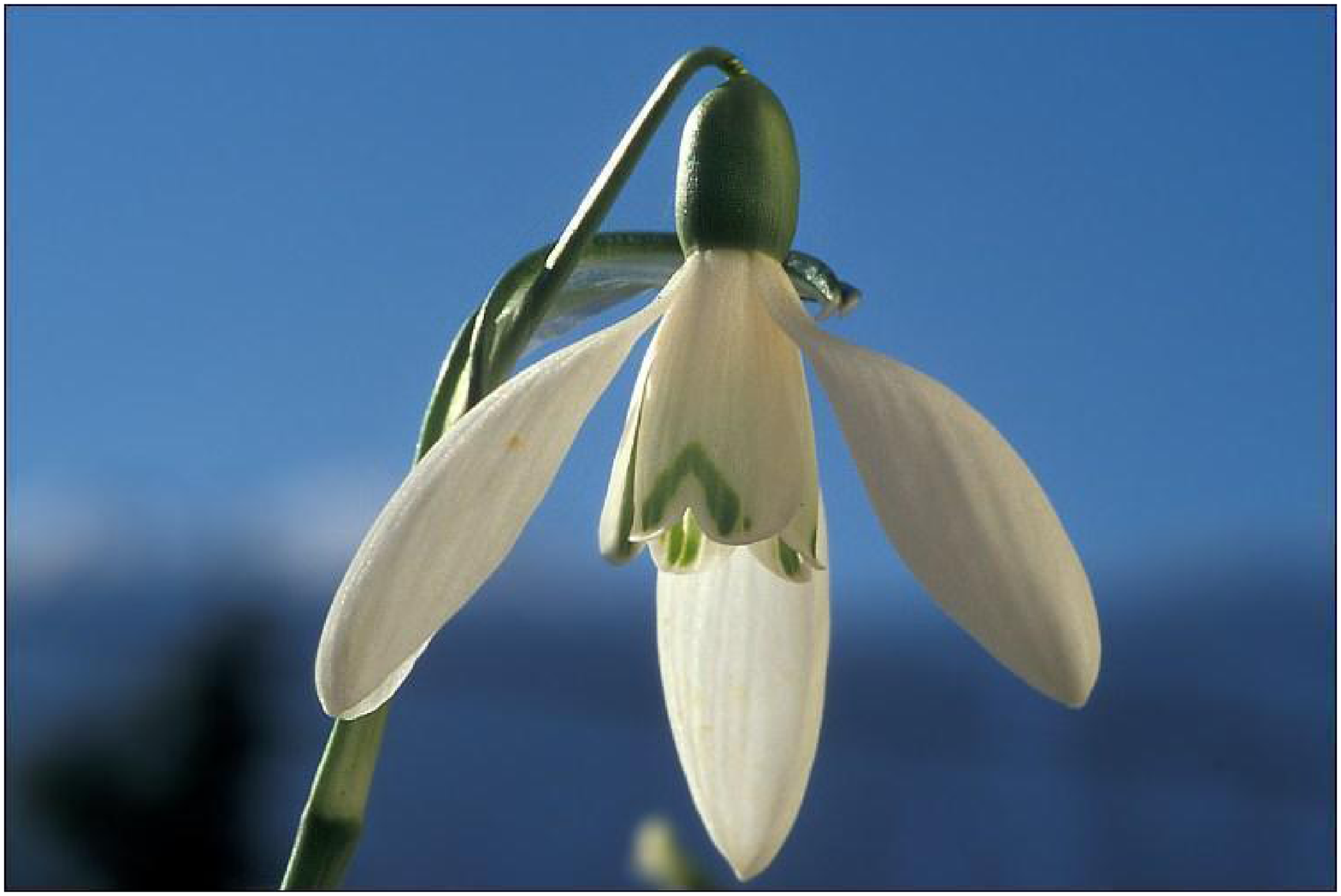} &
\includegraphics[width=0.23\linewidth]{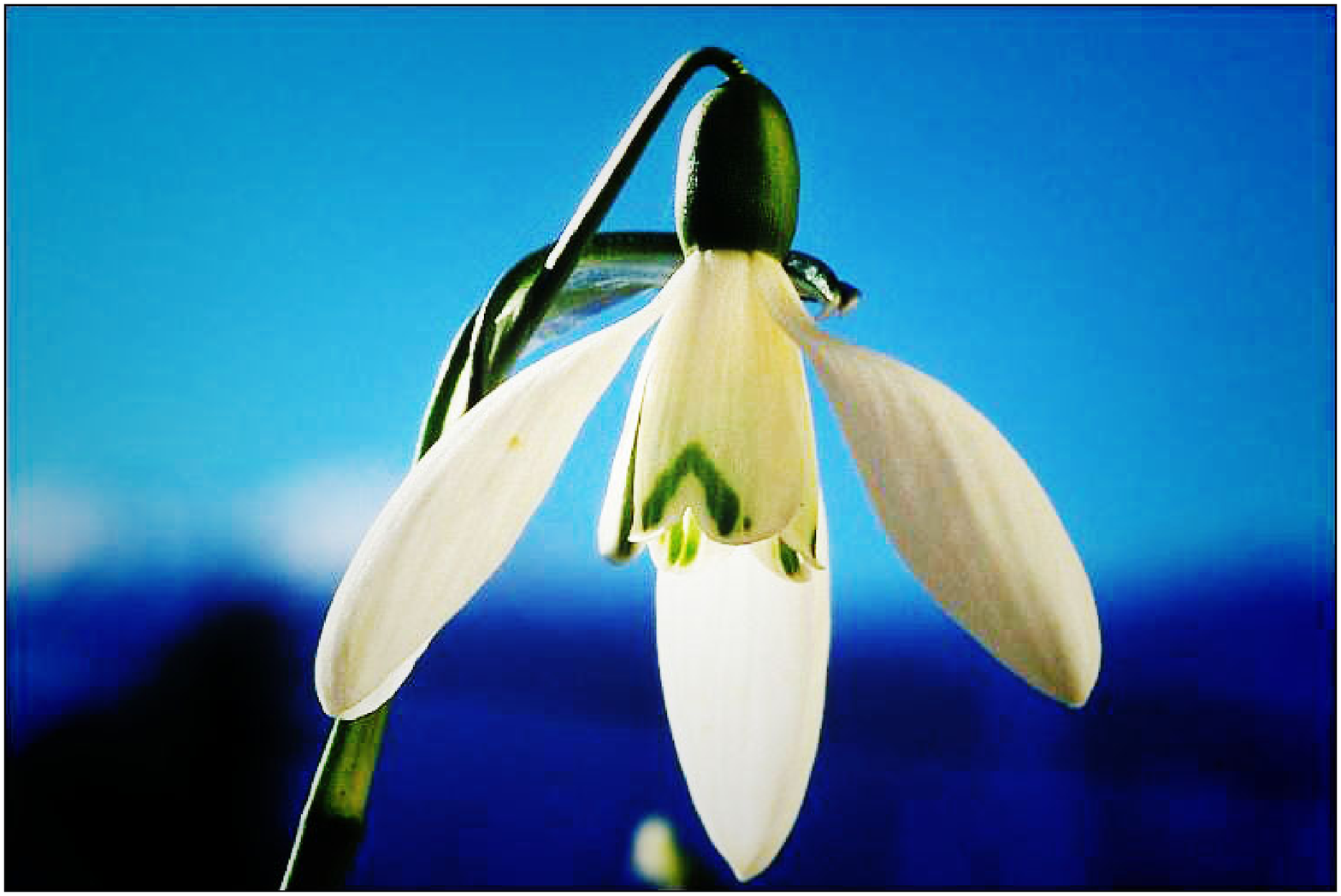} &\includegraphics[width=0.23\linewidth]{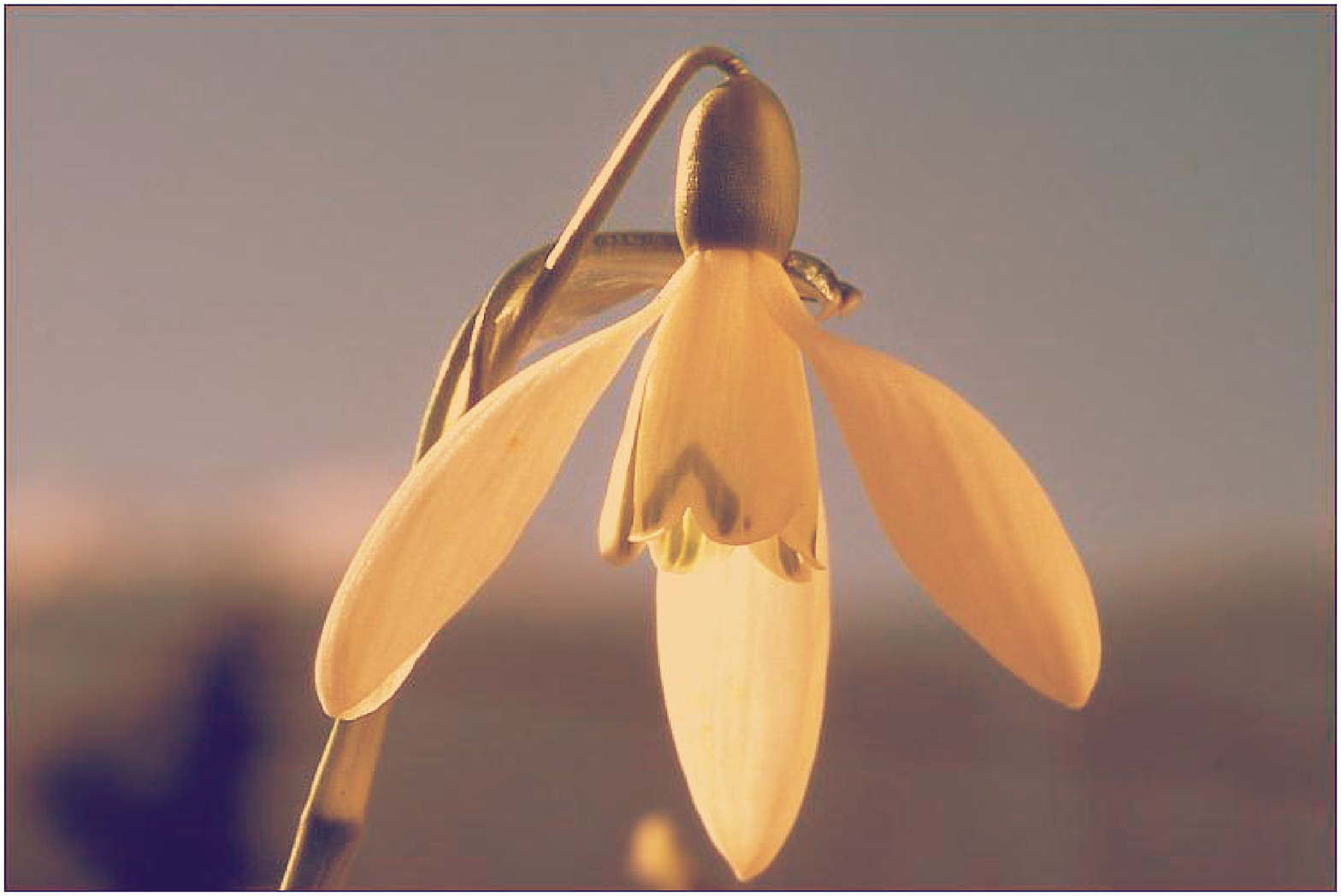} &\includegraphics[width=0.23\linewidth]{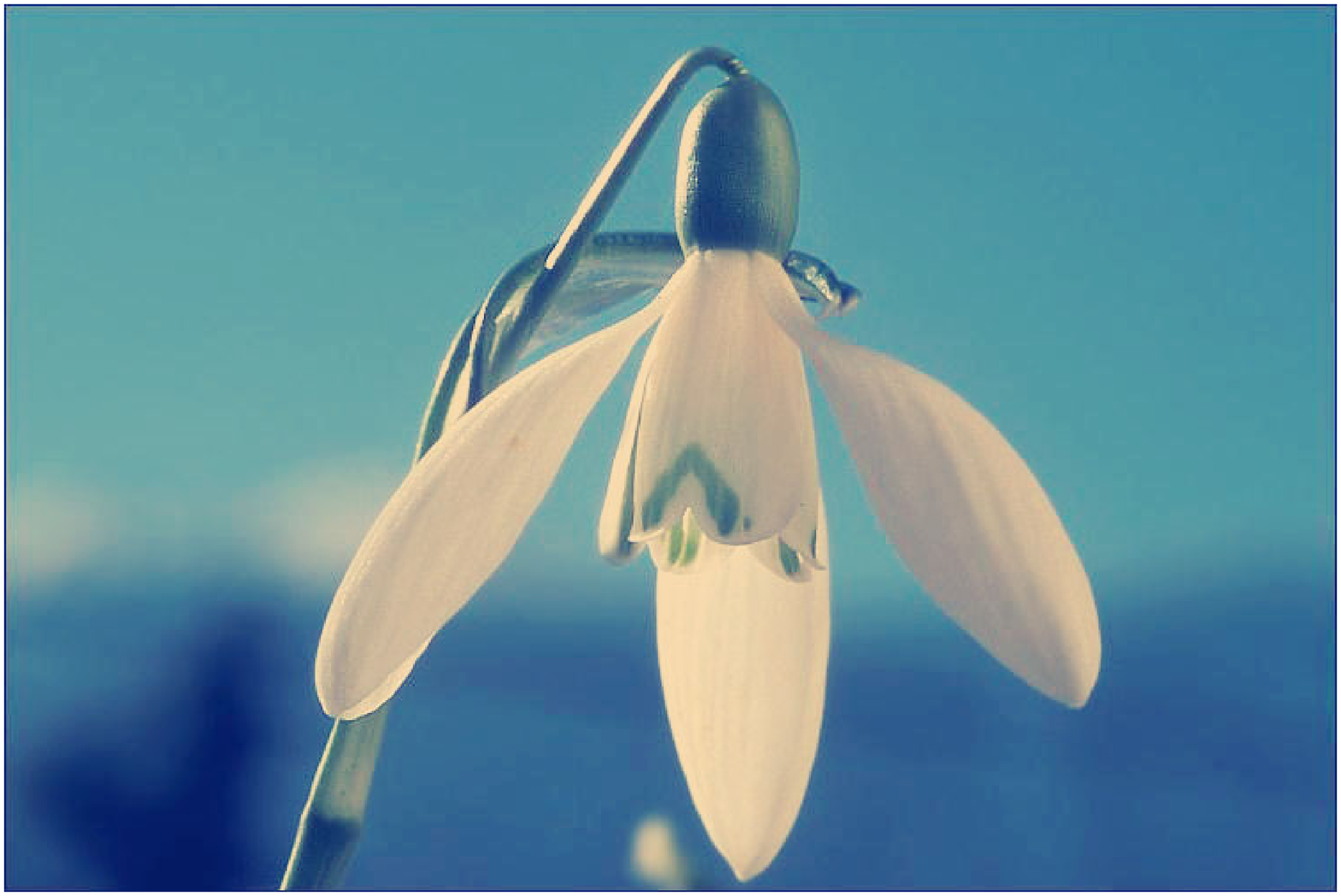}
\end{array}$
\end{center}
\caption{From left to right: original image, {\it lomo-fi}, {\it lord-kelvin}, and {\it Nashville}.}
\label{fig: flower_filters}
\end{figure*}

To simulate the images used in real world applications, which are taken by different devices and under through various of pixel-level editing, we process the image from Oxford Flower data set with 3 effect filters that are popularly used in Instagram\texttrademark: {\it lomo-fi}, {\it lord-kelvin}, and {\it Nashville}. Along with the original images, we obtain a image data set of 4 effects. Since the original flower images are collected from many difference sources, the original images were taken by different devices under different conditions, then the affects of scene radiance and image pipelines are already taken in to consideration. We show an example image and its 3 variants in Fig. \ref{fig: flower_filters}.

Since KDES gives a general framework for gradient-based descriptors, we only use KDES in the experiments in this section. For convenience of expression, we refer the data sets obtained by applying effect filters as picture {\it styles}. Similar as in Section \ref{sec: daadd}, we extract KDES for all the images from all 4 styles ({\it original, lomo-fi, lord-kelvin, and Nashville}). We construct a 2,000-word codebook by sampling 4 types of descriptors from {\it original} style. Then all the images from 4 effects are quantized using EMK with this codebook. To simulate the real image collections as mixtures of images with different picture styles, one image is only selected once from one of the 4 styles to obtain an experimental data set. For example, the 4 images in Fig. \ref{fig: flower_filters} are actually variants of the same image, then only one of them will be selected for one experiment. We conduct experiments on different style combinations to demonstrate the robustness of the proposed ADD method. In particular, experiments on single style data set are conducted for the 4 styles respectively. And the mixtures of the {\it original} and one of the other three styles are used. At last, a mixture of all 4 styles are generated. After constructing the experimental data sets uniformly, we randomly split the images as 40 per category for training and the rest for testing. Averagely speaking, equal numbers of images from different styles will appear in training and testing. Different from domain adaptation, images used here in training or testing are not separated by domain-ships, which is more similar to the real-world applications where there is no prior information about domain-ship or picture styles of the images. We perform object recognition using SVMs for the standard KDES, averaging kernel, and learned kernel by GMKL. We report the results for 10 runs of experimental data set generation and training/testing spliting for each scenario in Table\ref{tab: flower_kdes}.

\begin{table}
\begin{center}
\resizebox{1\linewidth}{!} {
\begin{tabular}{|l|l||c|c|c|}
\hline
style1 & style2  & standard KDES & ADD\_AK & ADD\_GMKL\\
\hline\hline
original & n/a  &    69.35 $\pm$ 2.20    &   67.76 $\pm$ 2.65 & {\bf 74.32} $\pm$ 1.77\\  
\hline 
original & lomo-fi   &   65.85 $\pm$ 1.66    &   64.24 $\pm$ 1.88 & {\bf 71.62} $\pm$ 1.04\\  
\hline 
original & lord-kelvin  &    67.53 $\pm$ 1.32    &   66.06 $\pm$ 1.80 & {\bf 72.82} $\pm$ 0.69\\
\hline  
original & Nashville  &    66.44 $\pm$ 1.73    &   64.62 $\pm$ 1.39 & {\bf 71.88} $\pm$ 0.72\\
\hline    
lomo-fi & n/a   &    65.12 $\pm$ 1.48    &   63.97 $\pm$ 1.62 & {\bf 69.82} $\pm$ 0.70 \\ 
\hline
lord-kelvin & n/a  &    68.09$\pm$ 1.38    &   67.06 $\pm$ 1.40 & {\bf 72.62} $\pm$ 1.38\\ 
\hline 
Nashville & n/a  &    67.03$\pm$ 1.10    &   66.85$\pm$ 1.28 & {\bf 71.68} $\pm$ 1.94\\  
\hline
 all & n/a     &    64.56 $\pm$ 0.90    &   63.24 $\pm$ 0.58 & {\bf 69.88} $\pm$ 0.51\\  
\hline
\end{tabular}
}
\end{center}
\caption{ Experiments on Oxford Flower data set. The average accuracy in \% is reported and the corresponding standard deviation is included.}
\label{tab: flower_kdes}
\end{table}

From Table\ref{tab: flower_kdes} we can clearly see that the proposed ADD\_GMKL method is superior than the standard KDES in all cases. From the top 4 rows of Table\ref{tab: flower_kdes}, we found that the recognition accuracies decrease when images are with different picture styles, which confirms the motivation we described in Section \ref{sec: intro}. In addition, according to the single style experimental results, pixel-level editing functions can influence the recognition accuracy through the computation of descriptors when the editing functions are applied to all the images used in training and testing, which supports the proposed idea that learning an optimal editing function $g^*$ can improve object recognition using gradient-based descriptors. In the last case, when images are uniformly sampled from all 4 styles, the standard KDES descriptor gives the worse performance, which is reasonable since the higher diversity of the images leads to larger differences between descriptors of similar image patches of the same objects.

After demonstrating the problem we introduced in this paper widely exists, the improved performance of ADD\_GMKL shows that our proposed algorithm is an efficient solution. Recall that our ADD can be considered as a single feature method, the proposed ADD\_GMKL outperforms the state-of-art\cite{gehler09feature} by 4\% on the original Oxford Flower data set. Therefore, the Adaptive Descriptor Design can be used widely on top of gradient-based descriptors to further improve the recognition accuracy.

We also notice that the ADD\_AK method is not better than the standard KDES. We believe it is caused by the small size of the codebook (2,000 words). Since 4 types of descriptors are extracted from one image on a dense grid and there are 1360 images in total, this codebook introduced large distortion in quantization, which decreases the performance of averaging kernel.

\section{Conclusion}
In this paper, we introduced a problem in object recognition caused by different picture styles of images. After showing the first study that links pixel-level editing functions with gradient-based image descriptors, we proposed an Adaptive Descriptor Design (ADD) to solve the problem. We demonstrated that ADD can be widely used as a general framework based on popular descriptors, and the experimental results show the improvements of ADD on domain adaptation data set, standard Oxford Flower data set and its variants with different picture styles.

{\small
\bibliographystyle{ieee}
\bibliography{bib_iccv}

\begin{thebibliography}{10}\itemsep=-1pt

\bibitem{bay08surf}
H.~Bay, A.~Ess, T.~Tuytelaars, and L.~Van~Gool.
\newblock Speeded-up robust features (surf).
\newblock {\em Computer vision and image understanding}, 110(3):346--359, 2008.

\bibitem{bo10kdes}
L.~Bo, X.~Ren, and D.~Fox.
\newblock Kernel descriptors for visual recognition.
\newblock {\em Advances in Neural Information Processing Systems (NIPS)}, 7,
  2010.

\bibitem{bo09efficient}
L.~Bo and C.~Sminchisescu.
\newblock Efficient match kernel between sets of features for visual
  recognition.
\newblock {\em Advances in neural information processing systems (NIPS)}, 2(3),
  2009.

\bibitem{dalal05hog}
N.~Dalal and B.~Triggs.
\newblock Histograms of oriented gradients for human detection.
\newblock In {\em IEEE Conference onComputer Vision and Pattern Recognition
  (CVPR)}, volume~1, pages 886--893. IEEE, 2005.

\bibitem{feifei06one}
L.~Fei-Fei, R.~Fergus, and P.~Perona.
\newblock One-shot learning of object categories.
\newblock {\em IEEE Transactions on Pattern Analysis and Machine Intelligence},
  28(4):594--611, 2006.

\bibitem{gehler09feature}
P.~Gehler and S.~Nowozin.
\newblock On feature combination for multiclass object classification.
\newblock In {\em IEEE International Conference on Computer Vision (ICCV)},
  pages 221--228. IEEE, 2009.

\bibitem{gong12geodesic}
B.~Gong, Y.~Shi, F.~Sha, and K.~Grauman.
\newblock Geodesic flow kernel for unsupervised domain adaptation.
\newblock In {\em IEEE Conference on Computer Vision and Pattern Recognition
  (CVPR)}, pages 2066--2073. IEEE, 2012.

\bibitem{gopalan11domain}
R.~Gopalan, R.~Li, and R.~Chellappa.
\newblock Domain adaptation for object recognition: An unsupervised approach.
\newblock In {\em IEEE International Conference on Computer Vision (ICCV)},
  pages 999--1006. IEEE, 2011.

\bibitem{grossberg04modeling}
M.~D. Grossberg and S.~K. Nayar.
\newblock Modeling the space of camera response functions.
\newblock {\em IEEE Transactions on Pattern Analysis and Machine Intelligence},
  26(10):1272--1282, 2004.

\bibitem{jhuo12robust}
I.-H. Jhuo, D.~Liu, D.~Lee, and S.-F. Chang.
\newblock Robust visual domain adaptation with low-rank reconstruction.
\newblock In {\em IEEE Conference on Computer Vision and Pattern Recognition
  (CVPR)}, pages 2168--2175. IEEE, 2012.

\bibitem{kim12new}
S.~Kim, H.~Lin, Z.~Lu, S.~Susstrunk, S.~Lin, and M.~Brown.
\newblock A new in-camera imaging model for color computer vision and its
  application.
\newblock {\em IEEE Transactions on Pattern Analysis and Machine Intelligence},
  2012.

\bibitem{kulis11art}
B.~Kulis, K.~Saenko, and T.~Darrell.
\newblock What you saw is not what you get: Domain adaptation using asymmetric
  kernel transforms.
\newblock In {\em IEEE Conference on Computer Vision and Pattern Recognition
  (CVPR)}, pages 1785--1792. IEEE, 2011.

\bibitem{lazebnik06spm}
S.~Lazebnik, C.~Schmid, and J.~Ponce.
\newblock Beyond bags of features: Spatial pyramid matching for recognizing
  natural scene categories.
\newblock In {\em IEEE Conference on Computer Vision and Pattern Recognition
  (CVPR)}, volume~2, pages 2169--2178. IEEE, 2006.

\bibitem{ling05deformation}
H.~Ling and D.~W. Jacobs.
\newblock Deformation invariant image matching.
\newblock In {\em IEEE International Conference on Computer Vision (ICCV)},
  volume~2, pages 1466--1473. IEEE, 2005.

\bibitem{lowe04sift}
D.~G. Lowe.
\newblock Distinctive image features from scale-invariant keypoints.
\newblock {\em International journal of computer vision}, 60(2):91--110, 2004.

\bibitem{moreno11deformation}
F.~Moreno-Noguer.
\newblock Deformation and illumination invariant feature point descriptor.
\newblock In {\em IEEE Conference on Computer Vision and Pattern Recognition
  (CVPR)}, pages 1593--1600. IEEE, 2011.

\bibitem{nilsback06flower}
M.-E. Nilsback and A.~Zisserman.
\newblock A visual vocabulary for flower classification.
\newblock In {\em IEEE Conference on Computer Vision and Pattern Recognition
  (CVPR)}, volume~2, pages 1447--1454. IEEE, 2006.

\bibitem{saenko10da}
K.~Saenko, B.~Kulis, M.~Fritz, and T.~Darrell.
\newblock Adapting visual category models to new domains.
\newblock In {\em European Conference on Computer Vision (ECCV)}, pages
  213--226. Springer, 2010.

\bibitem{simonyan12descriptor}
K.~Simonyan, A.~Vedaldi, and A.~Zisserman.
\newblock Descriptor learning using convex optimisation.
\newblock In {\em European Conference on Computer Vision （ECCV）}, pages
  243--256. Springer, 2012.

\bibitem{tola10daisy}
E.~Tola, V.~Lepetit, and P.~Fua.
\newblock Daisy: An efficient dense descriptor applied to wide-baseline stereo.
\newblock {\em IEEE Transactions on Pattern Analysis and Machine Intelligence},
  32(5):815--830, 2010.

\bibitem{varma09gmkl}
M.~Varma and B.~R. Babu.
\newblock More generality in efficient multiple kernel learning.
\newblock In {\em International Conference on Machine Learning (ICML)}, pages
  1065--1072. ACM, 2009.

\bibitem{wah11bird}
C.~Wah, S.~Branson, P.~Perona, and S.~Belongie.
\newblock Multiclass recognition and part localization with humans in the loop.
\newblock In {\em IEEE International Conference on Computer Vision (ICCV)},
  pages 2524--2531. IEEE, 2011.

\bibitem{winder09picking}
S.~Winder, G.~Hua, and M.~Brown.
\newblock Picking the best daisy.
\newblock In {\em IEEE Conference onComputer Vision and Pattern Recognition
  (CVPR)}, pages 178--185. IEEE, 2009.

\bibitem{winder07learning}
S.~A. Winder and M.~Brown.
\newblock Learning local image descriptors.
\newblock In {\em Computer Vision and Pattern Recognition, 2007. CVPR'07. IEEE
  Conference on}, pages 1--8. IEEE, 2007.

\bibitem{xiong12pixels}
Y.~Xiong, K.~Saenko, T.~Darrell, and T.~Zickler.
\newblock From pixels to physics: Probabilistic color de-rendering.
\newblock In {\em IEEE Conference on Computer Vision and Pattern Recognition
  (CVPR)}, pages 358--365. IEEE, 2012.

\end{thebibliography}
}

\end{document}